\documentclass[runningheads]{llncs}

\usepackage{eccv}

\usepackage{eccvabbrv}

\usepackage{graphicx}
\usepackage{booktabs}
\usepackage{soul}
\usepackage{multirow}
\usepackage{makecell}
\usepackage{svg}
\usepackage{tikz}
\usepackage{wrapfig}

\usepackage[accsupp]{axessibility}  
\usepackage{hyperref}

\usepackage{orcidlink}
\newcommand{\mypara}[1]{\noindent\textbf{#1.}~}

\newcommand{\cd}{\textrm{CD}}
\newcommand{\sdf}{\textrm{SDF}}
\newcommand{\normal}{\textrm{Norm}}

\newcommand{\smooth}{\textrm{Smo}}
\newcommand{\rcd}{\textrm{R-CD}}
\newcommand{\rsdf}{\textrm{R-SDF}}
\newcommand{\shape}{\textrm{Shape}}
\newcommand{\init}{\textrm{init}}

\begin{document}

% ---------------------------------------------------------------
\title{DynoSurf: Neural Deformation-based Temporally Consistent Dynamic Surface Reconstruction} 

\titlerunning{DynoSurf}

\author{Yuxin Yao\inst{1}\orcidlink{0000-0002-5410-0782} \and
Siyu Ren\inst{1}\orcidlink{0000-0001-8551-0787} \and
Junhui Hou\inst{1}\orcidlink{0000-0003-3431-2021}\thanks{Corresponding author. Email: \texttt{jh.hou@cityu.edu.hk}.}
\and
Zhi Deng\inst{2}
\and \\
Juyong Zhang\inst{3}\orcidlink{0000-0002-1805-1426}
\and
Wenping Wang\inst{4}
}

\authorrunning{Y.~Yao, S.~Ren et al.}

\institute{$^1$City University of Hong Kong \quad $^2$Tencent Games \\
$^3$University of Science and Technology of China
\quad $^4$Texas A\&M University
}
\maketitle

\begin{figure}[h] 
    \centering
    \includegraphics[width=0.9\columnwidth]{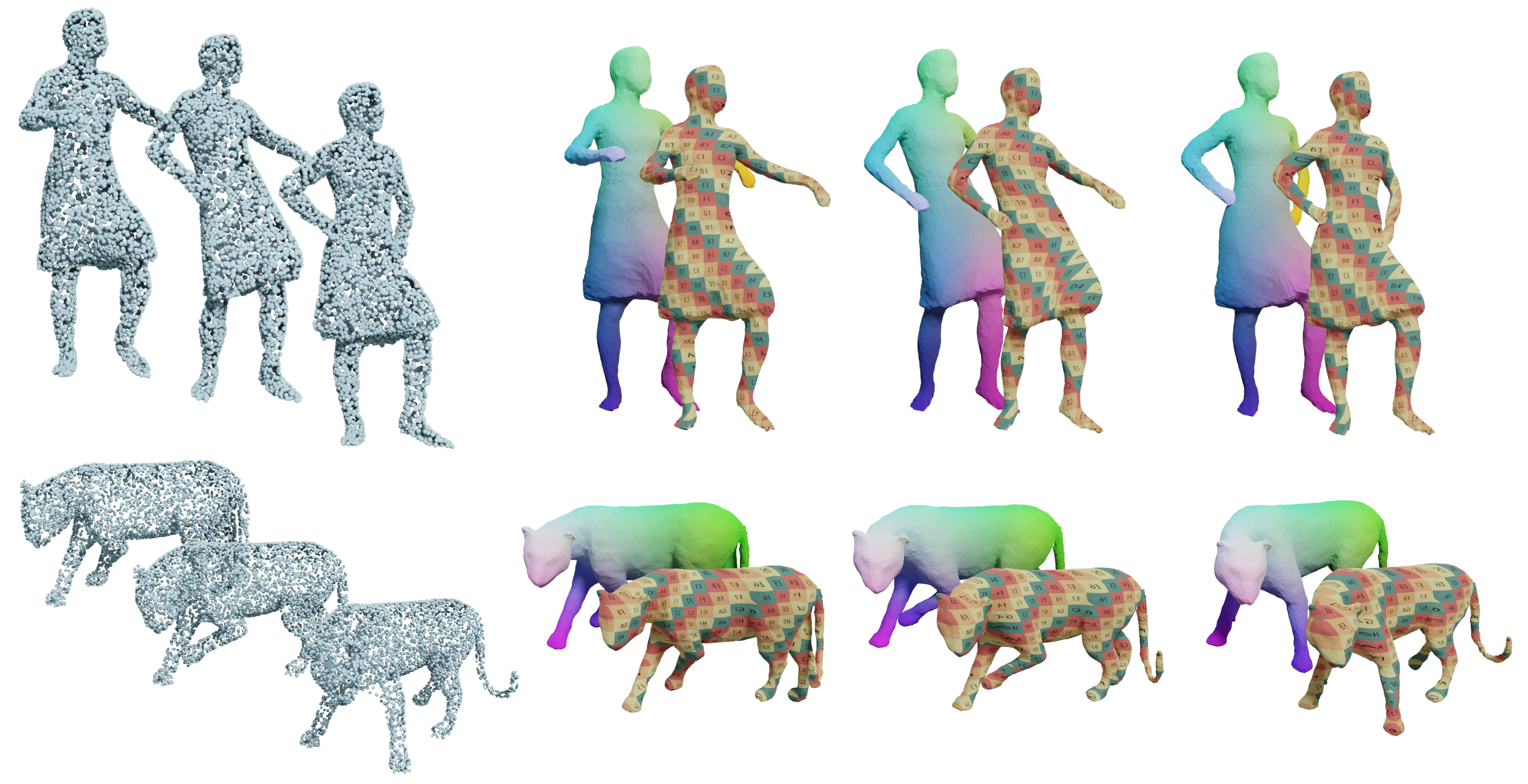}
    \caption{Temporally-consistent dynamic meshes (i.e., vertices are corresponded and connections are identical over time) reconstructed by our DynoSurf from {continuous dynamic} 3D point cloud sequences \textbf{without} using any shape-prior, ground-truth surface, and ground-truth temporal correspondence.   
    The color and texture map are used to illustrate correspondence across reconstructed mesh frames. 
    }
    \label{fig:demo} 
\end{figure}

\begin{abstract}
This paper explores the problem of reconstructing temporally consistent surfaces from a 3D point cloud sequence without correspondence. To address this challenging task, we propose DynoSurf, an unsupervised learning framework integrating a template surface representation with a learnable deformation field. Specifically, we design a coarse-to-fine strategy for learning the template surface based on the deformable tetrahedron representation. Furthermore, we propose a learnable deformation representation based on the learnable control points and blending weights, which can deform the template surface non-rigidly while maintaining the consistency of the local shape. 
Experimental results demonstrate the significant superiority of DynoSurf over current state-of-the-art approaches, showcasing its potential as a powerful tool for dynamic mesh reconstruction. The code is publicly available at \url{https://github.com/yaoyx689/DynoSurf}.
\end{abstract}

\section{Introduction} 

The reconstruction of temporally consistent geometric surfaces from a continuous dynamic 3D point cloud sequence without correspondences presents a valuable yet formidable challenge. These geometric surfaces play a crucial role in diverse domains, including film, gaming, animation, virtual reality (VR), augmented reality (AR), and robotics. They facilitate various applications such as motion object editing, texture transfer, and shape analysis, thereby contributing to the creation of realistic visual effects and immersive experiences.

With the development of the static surface reconstruction for a single object from a point cloud~\cite{kazhdan2006poisson,shen2021deep,peng2021shape,huang2023neural}, 
a natural idea to model the temporally consistent dynamic surfaces is to select a keyframe from the point cloud sequences and reconstruct it as a template surface through static reconstruction technology. 
Then the non-rigid deformation for the template surface is estimated by aligning it with other frames.
High requirements are placed on the selection of the keyframe point cloud with fewer missing parts or occlusions and a similar shape to other frames. 
Some optimization-based dynamic reconstruction frameworks~\cite{newcombe2015dynamicfusion,dou2016fusion4d,slavcheva2017killingfusion} continuously update the template shape during frame-by-frame non-rigid tracking. These approaches are effective and pragmatic, but most of them use depth as input and cannot handle the sparse point clouds without any camera information. 
Some learning-based method~\cite{niemeyer2019occupancy,tang2021learning,lei2022cadex}
no longer define or pre-establish the template shape, but jointly learn dynamic reconstruction. This way may achieve better results but increases the difficulty of decoupling the deformation field and the template space, which makes deformation have less practical significance. 
Moreover, these methods use ground truth to supervise training, have poor generalization, and can usually only process sequences of fixed length.
To address these problems, we first design an adaptive keyframe selection strategy and learn the template surface representation, then combine it with a deformation field to form a dynamic reconstruction framework.

Furthermore, the design of the deformation field is particularly important. Optimization-based methods~\cite{li2008global,li2009robust,yao2020quasi} often employ the embedded deformation graph~\cite{sumner2007embedded} for non-rigid deformation, but they depend on the spatial position and mesh quality of the template surface. Some methods~\cite{zhang2023self,park2021nerfies,tretschk2021non} directly use a network to predict the deformation for each point, lacking explicit control of the local structure. 
\cite{bozic2021neural} proposes a neural deformation graph to reconstruct dynamic objects. However, they have some restrictions on the location and connection relationships of nodes, which necessitates the use of multiple loss functions. Therefore, we propose a deformation field based on explicit learnable control points, which improves the explicit controllability of the deformation field based on implicit representation. We also learn the blending weights between the points on the template surface and the control points. This allows us to adaptively learn the deformation consistency of surface points based on motion information.

In this paper, we aim to reconstruct a temporally consistent surface from continuous dynamic 3D point cloud sequences with not-fixed length. We propose a reconstruction framework combining a template surface representation and a learnable deformation field based on the learnable control points and blending weight. 
Extensive experiments on three benchmark datasets demonstrate that our proposed temporally-consistent surface reconstruction outperforms the state-of-the-art methods to a large extent. 

To summarize, our contributions are as follows: 
\begin{itemize}
    \item We propose a new learning framework for reconstructing temporally-consistent dynamic surfaces from {continuous dynamic} point cloud sequences that can be trained  \textit{without} requiring any shape prior,
    ground-truth surface, and ground-truth temporal correspondence information as supervision.
 
    \item We propose a coarse-to-fine learning strategy for unsupervised static surface reconstruction based on the deformable tetrahedron representation. 

    \item We propose a deformation representation based on learnable control points and blending weights that can better maintain local structure while being independent of object category.
    
\end{itemize}
\section{Related Work} 

We briefly review the most relevant works, 
including 3D shape representation, non-rigid shape deformation, and sequential shape reconstruction.

\mypara{3D Shape Representation}
Generally, 3D shape representation could be divided into two main categories, explicit and implicit, and these can be transformed into one another. The point cloud and triangle mesh are the two most commonly utilized explicit representations for 3D shapes in various applications \cite{qi2017pointnet, li2018pointcnn, zeng20173dmatch, hanocka2019meshcnn, feng2019meshnet}. 
Implicit representation utilizes an isosurface of a field to depict a surface. The Binary Occupancy Field (BOF) \cite{kazhdan2013screened, mescheder2019occupancy, peng2020convolutional, boulch2022poco} and Signed Distance Field (SDF) \cite{kolluri2008provably, liu2021deep, park2019deepsdf} are two prevalent implicit fields for representing shapes, both dividing the entire space into two regions, inside and outside the shape. The implicit fields could be approximated by a neural network, such as DeepSDF \cite{park2019deepsdf}, and SAL \cite{atzmon2020sal}.

Given the shapes explicitly represented as point clouds or triangle meshes, diverse methods exist for their conversion into implicit representations \cite{niemeyer2019occupancy, kazhdan2013screened, kolluri2008provably, liu2021deep, koneputugodage2023octree}. Conversely, when transforming implicit shapes into explicit forms, Marching Cubes (MC) \cite{MARCHINGCUBE} and Marching Tetrahedrons (MT) \cite{nielson2008dual} stand out as the two widely used methods. Considering the detailed parts of shapes, DMTet \cite{shen2021deep} and FlexiCubes \cite{shen2023flexible} introduce offset for each tetrahedron or cube vertex, allowing their methodologies to effectively manage detailed shape components.

\mypara{Non-rigid Shape Deformation}
Deforming non-rigid shapes presents a challenge due to the high degree of freedom involved in controlling the shape deformation. To mitigate this, \cite{li2008global,li2009robust} employ an embedded deformation graph~\cite{sumner2007embedded} with local rigid properties for modeling deformation. During optimization, various geometric regularization terms, such as as-conformal-as-possible deformations \cite{wu2019global} and as-rigid-as-possible deformations \cite{yang2019global}, are incorporated to impart meaningful deformation. Recently,  \cite{yao2020quasi,yao2023fast} have developed optimization methods to accelerate the optimization process.
With the advancement of deep learning in 3D vision, numerous learning-based deformation methods have been introduced.  
CorrNet3D \cite{zeng2021corrnet3d} designs a symmetric pipeline to compute deformations between point clouds. RMA-Net \cite{feng2021recurrent} introduces a new deformation representation with a point-wise combination of several rigid transformations. 
NDP~\cite{li2022non} proposes a non-rigid motion representation using a pyramid architecture.

\mypara{Sequential Shape Reconstruction}
Several 3D parametric models, such as SMPL~\cite{SMPL2015} and SMAL~\cite{zuffi20173d}, have been developed to represent general shapes within specific categories. These parametric models allow for straightforward adjustments in pose and shape by modifying their respective parameters. Subsequently, the combination of neural implicit functions and parametric models enhances the expressive power for reconstruction~\cite{palafox2021npms,jiang2022h4d,jiang2022lord,chen2024neural}. However, these methods are constrained to handling sequences within specific categories, limiting their broad applicability.

Recent studies \cite{newcombe2015dynamicfusion, niemeyer2019occupancy, tang2021learning, bozic2021neural} have expanded the scope of reconstructing dynamic shapes by predicting deformation fields between different frames. Approaches like Dynamic Fusion \cite{newcombe2015dynamicfusion} and NDG \cite{bozic2021neural} establish a deformation graph to capture the deformation fields. O-Flow \cite{niemeyer2019occupancy} extends ONet \cite{mescheder2019occupancy} from 3D to 4D, representing deformation fields with a Neural-ODE \cite{chen2018neural}, while LPDC \cite{tang2021learning} employs MLPs to model correspondences in parallel. Both O-Flow and LPDC reconstruct the reference shape based on the initial frame, with subsequent frames having their query points mapped to the chosen reference frame. In contrast, I3DMM \cite{yenamandra2021i3dmm} learns a canonical reference shape for sequences of heads, while CASPR \cite{rempe2020caspr} and Garment Nets \cite{chi2021garmentnets} use ground truth canonical coordinates for network training, which is often impractical. Cadex \cite{lei2022cadex} employs canonical map factorization to represent deformation without requiring correspondences during training, but it may struggle with sequences featuring substantial motions. Recently, FDN~\cite{baieri2023fluidARXIV} proposes a new computational
model that leverages fluid simulation priors to handle dynamic 3D scenes with topological changes.
\section{Proposed Method}
\subsection{Problem Statement and Method Overview} 
Let $\{\mathbf{P}_1, ..., \mathbf{P}_K\}$ be a continuous dynamic 3D point cloud sequence capturing the geometric changes or motion of
scenes/objects over time, where the correspondence across frames is unknown and $\mathbf{P}_k\in\mathbb{R}^{N_k\times 3}$ is the $k$-th ($k\in[1,~K]$) frame containing $N_k$ points. We aim to reconstruct from them a \textit{temporally-consistent} mesh sequence 
$\{\mathcal{M}_1, ...,\mathcal{M}_k\}$
where the $k$-th mesh surface $\mathcal{M}_k:=\{\mathbf{V}_k, \mathbf{F}\}$ with $\mathbf{V}_k$ and $\mathbf{F}$ being the vertex set and face set, respectively,   
the vertices across meshes are corresponded, and all meshes share the same $\mathbf{F}$.

\begin{figure}[h]
    \centering
    \includegraphics[width=0.9\columnwidth]{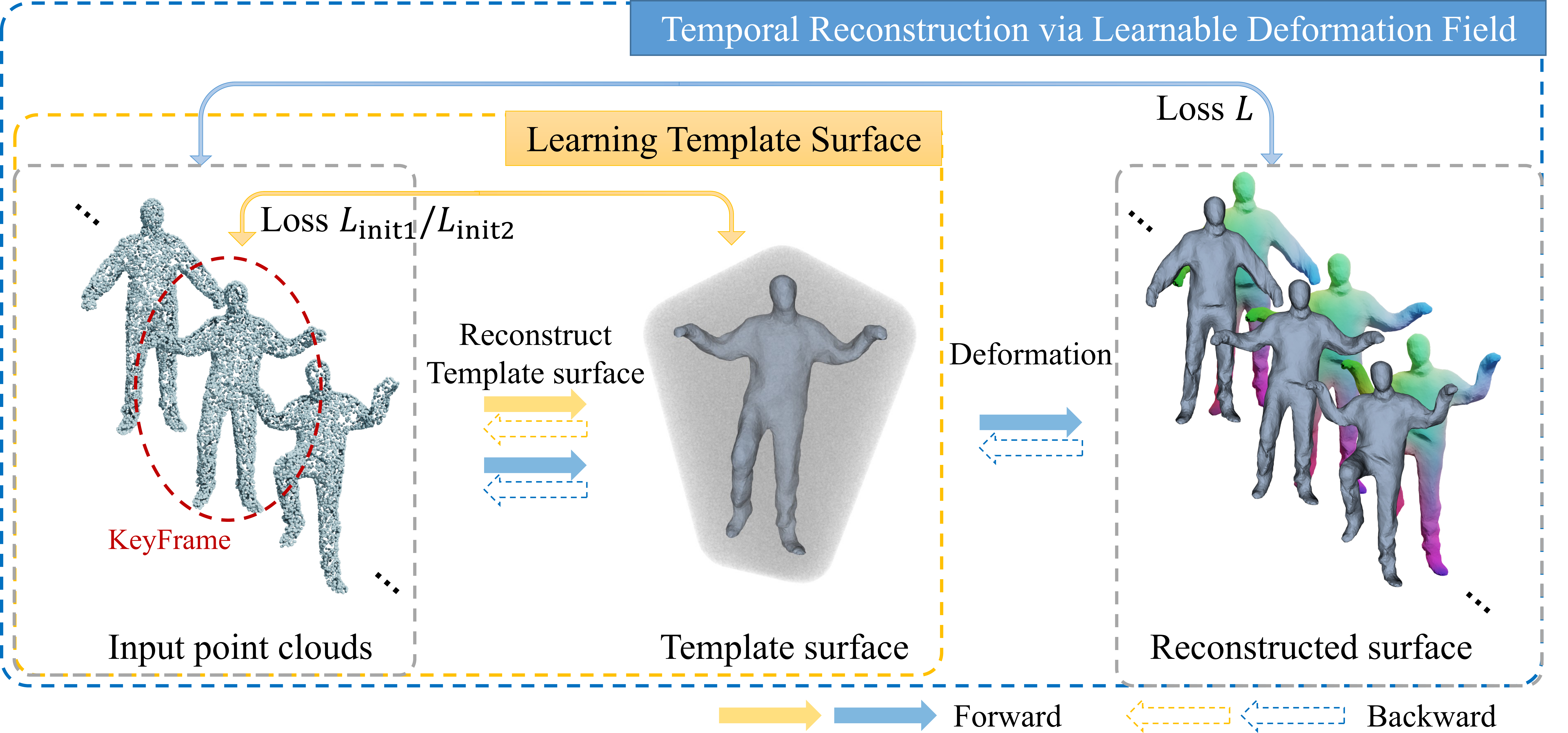} 
    \caption{Illustration of the proposed DynoSurf, which can reconstruct from continuous dynamic point cloud sequences temporally-consistent dynamic surfaces without requiring any ground-truth surface and temporal correspondence information. 
    }
     \label{fig:flowchart} 
\end{figure}

To address this challenging task, we propose an \textit{unsupervised} learning framework, named DynoSurf, which requires no ground-truth surface and correspondence information as supervision. As illustrated in Fig. \ref{fig:flowchart},   
DynoSurf mainly consists of two stages: template surface construction (Sec.~\ref{sec:shape-module}) and deformation-based temporal reconstruction (Sec.~\ref{sec:deformation-module}).  
Generally, we first select a keyframe from the input point clouds, denoted as $\mathbf{P}_{k^*}$, and reconstruct its surface based on the deformable tetrahedron representation through a coarse-to-fine learning process. Then, we deform the enhanced template surface to simultaneously align 
with all point clouds by learning a control points blending-based deformation field through jointly optimizing the template surface and deformation field. In what follows, we will detail the framework.

\subsection{Template Surface Representation via Deformable Tetrahedron} 
\label{sec:shape-module}
In this stage, we aim to construct a template surface, i.e., a mesh fitting the selected keyframe point cloud. 
As illustrated in Fig. \ref{fig:flowchart_template}, we initially select a keyframe $\mathbf{P}_{k^*}$, and construct its convex hull as a triangle mesh, referred to as $\mathcal{C}$. Subsequently, we process $\mathcal{C}$ to generate a tetrahedron mesh, denoted as $\mathcal{T}$, where the signed distance field (SDF) of the surface underlying  $\mathbf{P}_{k^*}$ is defined and learned through a coarse-to-fine 
learning process. The SDF enables us to extract a mesh, denoted as $\overline{\mathcal{M}}:=\{\overline{\mathbf{V}}, \mathbf{F}\}$ with $\overline{\mathbf{V}}$ being the vertex set. 

\input{figs_scripts/flowchart_template}

\mypara{Keyframe selection and tetrahedron mesh construction}
Since we will deform the reconstructed template surface from $\mathbf{P}_{k^*}$ for temporal reconstruction, 
intuitively,  
$\mathbf{P}_{k^*}$ should be relatively average to facilitate the deformation process. 
Thus, we determine 
$k^* = \mathop{\arg\min}_{k\in[1,K]} \sum_l \texttt{CD}_{\ell_2}(\mathbf{P}_k, \mathbf{P}_l)$,
where $\texttt{CD}_{\ell_2}(\cdot,~\cdot)$ computes the $\ell_2$-norm-based Chamfer distance between two point clouds. 

We then adopt QuickHull~\cite{barber1996quickhull} to construct the convex hull $\mathcal{C}$ of $\mathbf{P}_{k^*}$,  
which is further dilated and re-meshed to obtain a uniform triangle mesh $\widehat{\mathcal{C}}$ surrounding $\mathcal{C}$.  
By tetrahedralizing $\widehat{\mathcal{C}}$, we obtain a tetrahedron mesh $\mathcal{T}=\{\mathbf{Q}, \mathbf{T}\}$ with 
$\mathbf{Q}$ and $\mathbf{T}$ being the vertex and tetrahedron sets, respectively. 
Note that the vertices of $\mathbf{Q}$ are distributed both inside and outside of $\mathcal{C}$ due to the dilation and remeshing operations. 

\mypara{Coarse-to-fine learning of deformable tetrahedron-based template surface}  
We learn the deformable tetrahedron representation~\cite{shen2021deep} 
to construct a template surface $\overline{\mathcal{M}}$ that approximates the point cloud $\mathbf{P}_{k^*}$. 
\begin{equation}
[s(\mathbf{q}), \delta(\mathbf{q})] = F_{\bf{\Phi}}(\Gamma(\mathbf{q})), 
\end{equation} 
where $\Gamma(\cdot)$ is the Position Encoding operator. 
Then, based on the deformed tetrahedral grid points $\{\mathbf{q}+\delta(\mathbf{q})\}_{\mathbf{q}\in\mathbf{Q}}$ and  SDF values $\{s(\mathbf{q})\}_{\mathbf{q}\in\mathbf{Q}}$, we can employ the differentiable marching tetrahedron algorithm~\cite{shen2021deep}, denoted as $\texttt{DMT}(\cdot, ~\cdot)$, to extract $\overline{\mathcal{M}}$, i.e.,  
$\overline{\mathcal{M}}=\texttt{DMT}(\{\mathbf{q}+\delta(\mathbf{q})\}, \{s(\mathbf{q})\})$.

To train $F_{\bf{\Phi}}(\cdot)$ to encode the SDF of the surface  $\overline{\mathcal{M}}$, we adopt the following coarse-to-fine 
strategy to make the learning process easier and faster:
(\textbf{1}) We initialize $F_{\bf{\Phi}}(\cdot)$ by learning the SDF of $\mathcal{C}$, i.e.,  
\begin{equation}
\label{eq:init-step1}
\min_{\bf{\Phi}}L_{\init1} = \sum_{\mathbf{q}\in\mathbf{Q}} \|s(\mathbf{q})-\texttt{SDF}(\mathbf{q}, \mathcal{C})\|^2, 
\end{equation}
where $s(\mathbf{q})$ is the predicted SDF value of \texttt{$F_{\bf{\Phi}}(\cdot)$} and $\texttt{SDF}(\mathbf{q}, \mathcal{C})$ denotes the ground-truth SDF value of convex  hull surface $\mathcal{C}$ at $\mathbf{q}$. 
\noindent (\textbf{2}) We then refine $F_{\bf{\Phi}}(\cdot)$ by constraining $\overline{\mathcal{M}}$ and
$\mathbf{P}_{k^*}$ to be as close as possible, i.e., 
\begin{equation}
\label{eq:init-step2}
\min_{\bf{\Phi}}L_{\init2} = \tilde{w}_1 L_{\cd} + \tilde{w}_2 L_{\normal}^* + \tilde{w}_{3} L_{\sdf}, 
\end{equation}
where $L_{\cd}:=\texttt{CD}_{\ell_1}(\overline{\mathcal{M}}, \mathbf{P}_{k^*})$ denotes the $\ell_1$-norm-based Chamfer distance between $\overline{\mathcal{M}}$ and $\mathbf{P}_{k^*}$; $L_{\normal}^*:=\texttt{NC}_{\ell_1}(\overline{\mathcal{M}}, \mathbf{P}_{k^*})$ measures the consistency of the corresponding normals (see the \textit{Supplementary Material} for details);
$L_{\sdf}$ is employed to promote the accuracy of the learned SDF, defined as 
\begin{equation}
L_{\sdf} = \frac{1}{|\mathbf{Q}|}\sum_{\mathbf{q}\in\mathbf{Q}} |s(\mathbf{q}) - \texttt{SDF}_{{\rm IMLS}}(\mathbf{q}, \mathbf{P}_{k^*})|, 
\end{equation}
where $\texttt{SDF}_{{\rm IMLS}}(\mathbf{q}, \mathbf{P}_{k^*})$ is the SDF value at $\mathbf{q}$ approximated through implicit moving least-squares \cite{kolluri2008provably} 
(refer to the \textit{Supplementary Material}).  

\mypara{Remark} 
It is worth noting that the resulting template surface $\overline{\mathcal{M}}$ will be \textit{adaptively} updated/enhanced during the following deformation stage to facilitate the deformation stage,  thus promoting the quality of reconstructed surfaces of the remaining point cloud frames.

\subsection{Temporal Reconstruction via Learnable Deformation Field} 
\begin{figure}
    \centering
    \includegraphics[width=0.9\columnwidth]{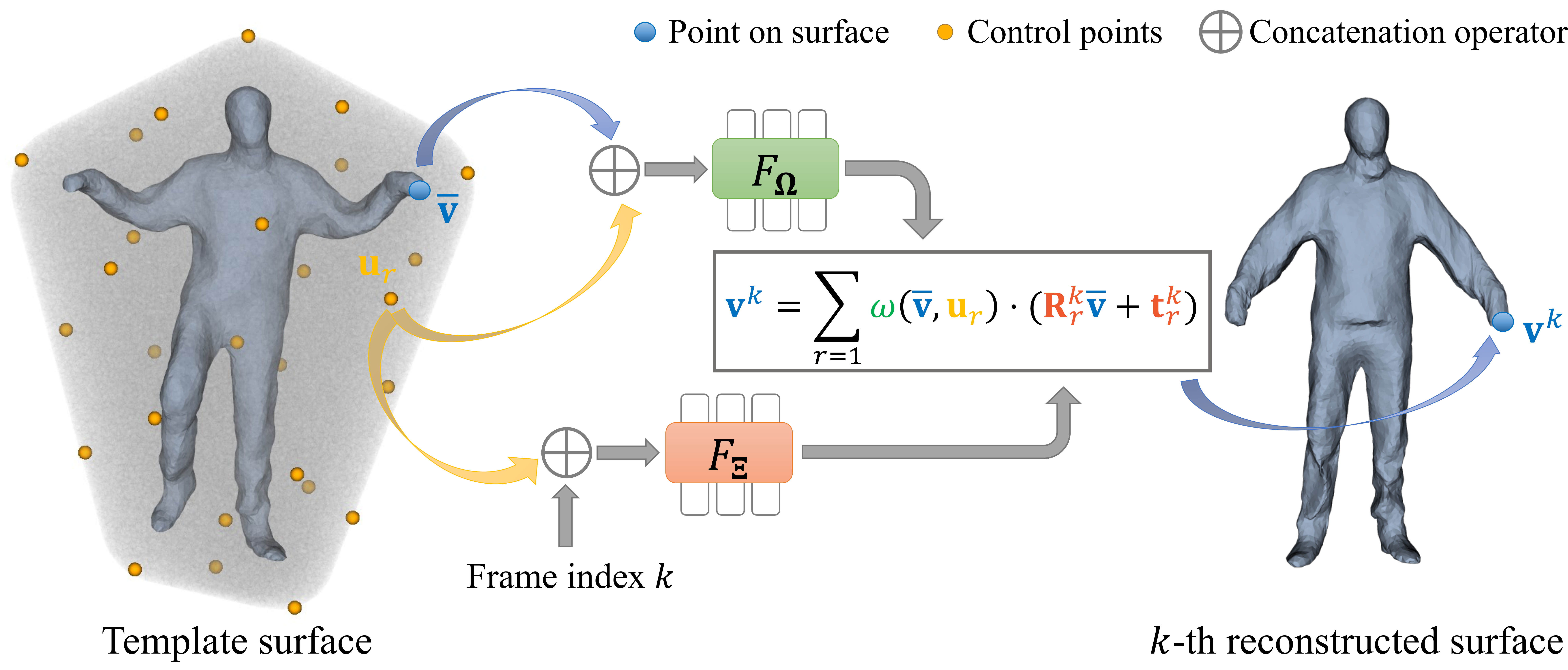} 
    \caption{Illustration of the proposed control points blending-based learnable deformation stage for temporal reconstruction. Note that the adaptively enhanced template surface will be deformed to all frames.}
    \label{fig:flowchart_deformation} 
\end{figure}

In this stage, as illustrated in Fig. \ref{fig:flowchart_deformation},  we propose a learnable deformation field based on control points blending. This deformation field enables us to deform the \textit{adaptively-enhanced} template surface to achieve temporal reconstruction. 
By employing this approach, we ensure the temporal consistency of the reconstructed surface sequence. 

\mypara{Control points blending-based learnable deformation field}
\label{sec:deformation-module} 
We first build a set of control point $\mathbf{U}:=\{\mathbf{u}_1,...,\mathbf{u}_{|\mathbf{U}|}\}$, which are obtained using the farthest point sampling (FPS) algorithm~\cite{eldar1997farthest} on the vertex set of the tetrahedron mesh $\mathcal{T}$ (i.e., $\mathbf{Q}$)  such that the control points $\mathbf{u}_r$, $r=1,2,\dots, |\mathbf{U}|$ are uniformly distributed in $\mathcal{T}$. Note that the positions of initial control points are also adaptively optimized during training to make the deformation more flexible. We associate each $\mathbf{u}_r$ with a rotation matrix $\mathbf{R}_r^k\in \texttt{SO}(3)$ and a translation $\mathbf{t}_r^k\in\mathbb{R}^{3}$ for each frame $\mathbf{P}_{k}$, $k=1,2,\dots, K$.  
Then for each vertex of the template surface   $\mathbf{\overline{v}}\in\overline{\mathbf{V}}$, which is the vertex set of the surface $\overline{\mathcal{M}}$. The deformed vertex at the $k$-th frame is represented as 
\begin{equation}
\label{eq:deformation}
\mathbf{v}^k = \sum_{r=1}^{|\mathbf{U}|} \omega(\mathbf{\overline{v}}, \mathbf{u}_r)\cdot(\mathbf{R}_r^k\mathbf{\overline{v}} + \mathbf{t}_r^k), 
\end{equation}
with the weight $\omega(\mathbf{\overline{v}}, \mathbf{u}_r)$ learned via 
\begin{equation}
\label{eq:learnable weights}
\omega(\mathbf{\overline{v}},\mathbf{u}_r) = \frac{\exp (F_{\bf{\Omega}}([\mathbf{\overline{v}}||(\mathbf{\overline{v}}-\mathbf{u}_r)]))}{\sum_{r=1}^{|\mathbf{U}|}\exp(F_{\bf{\Omega}}([\mathbf{\overline{v}}||(\mathbf{\overline{v}}-\mathbf{u}_r)]))}, 
\end{equation}
where ${F}_{\bf{\Omega}}(\cdot)$ is an MLP parameterized with $\bf{\Omega}$, and $[\cdot||\cdot]$ is the concatenation operator. 
Before training, we pre-optimize $\bf{\Omega}$ so that $\omega(\mathbf{\overline{v}},\mathbf{u}_r)=\tilde{\omega}(\mathbf{\overline{v}},\mathbf{u}_r)$, where
\begin{equation}
\label{eq:init_blending_weight}
\tilde{\omega}(\mathbf{\overline{v}},\mathbf{u}_r)=\frac{\exp(-\|\mathbf{\overline{v}}-\mathbf{u}_r\|^2/2\eta^2)}{\sum_{r=1}^{|\mathbf{U}|}\exp(-\|\mathbf{\overline{v}}-\mathbf{u}_r\|^2/2\eta^2)}, 
\end{equation}
The reconstructed surface of $\mathbf{P}_k$ can be obtained as $\mathcal{M}_k=\{\mathbf{V}_k, \mathbf{F}\}$ with $\mathbf{V}_k$ being the collection of $\mathbf{v}^k$.

We also utilize an MLP $F_{\bf{\Xi}}(\cdot)$ parametrized with $\bf{\Xi}$ to learn the deformation transformations of all frames, i.e.,  
$
[\xi_r^k,\mathbf{t}_r^k] = F_{\bf{\Xi}}(\Gamma([\mathbf{u}_r||k]))  
$, 
where $\xi_r^k\in \texttt{so}(3)$ is the Lie algebra format of $\mathbf{R}_r^k$. Owing to the spectral bias property of neural networks \cite{rahaman2019spectral},  
the learned transformations would vary smoothly in both spatial and temporal domains to ensure the spatial and temporal smoothness of the constructed mesh sequence.

\mypara{Learning of deformable field with enhanced template surface}
Instead of solely focusing on training the deformation stage, i.e., $F_{\bf{\Omega}}(\cdot)$, $F_{\bf{\Xi}}(\cdot)$ and $\mathbf{U}$, we jointly train the deformation stage and the template surface representation, i.e., $F_{\bf{\Phi}}(\cdot)$ has been warmed-up in the first stage. Specifically, at each iteration, we extract the template surface using $\texttt{DMT}(\cdot, \cdot)$ and then deform it to all frames for temporally consistent reconstruction.
Through joint optimization, we aim to enhance the adaptability of the template surface to the deformation process, ultimately leading to improved temporal reconstruction quality.  To achieve this goal, we consider the following loss function: 
\begin{equation}
\min_{\bf{\Phi}, \bf{\Omega}, \bf{\Xi}, \mathbf{U}} L = w_{1} L_{\rcd} + w_{2} L_{\normal} + w_{3} L_{\rsdf} + w_{4} L_{\smooth} + w_{5} L_{\shape}. 
\end{equation}

Specifically, $L_{\rcd}$ stands for the proposed robust Chamfer distance to cope with point clouds with holes and outliers, defined as
$L_{\rcd} = \frac{1}{K}\sum_{k} L_{\rcd}^k$
with 
\begin{equation}
\label{eq:robust_chamfer}
 L_{\rcd}^k = \frac{1}{|\widetilde{\mathbf{V}}_k|}\sum_{\mathbf{v}_i\in\widetilde{\mathbf{V}}_k} \phi_{\alpha}({\mathbf{v}}_i,\mathbf{p}_{\rho_i})\cdot\|{\mathbf{v}}_i - \mathbf{p}_{\rho_i}\|^2  + \frac{1}{|\mathbf{P}_k|}\sum_{\mathbf{p}_j\in\mathbf{P}_k} \phi_{\alpha}(\mathbf{p}_j,{\mathbf{v}}_{\tau_j})\cdot\|\mathbf{p}_j-{\mathbf{v}}_{\tau_j}\|^2, 
\end{equation}
where $\widetilde{\mathbf{V}}_k$ is the set of points uniformly sampled from $\mathcal{M}_k$, and $\phi_{\alpha}(x,y) = \exp(-\alpha \cdot \|x-y\|^2)$.  
$\mathbf{p}_{\rho_i}$ and $\mathbf{v}_{\tau_j}$ are the closest points for $\mathbf{v}_i$ and $\mathbf{p}_j$ respectively. 
The normal loss term $L_{\normal}=\sum_{k}\texttt{NC}_{\ell_1}(\widetilde{\mathbf{V}}_k, \mathbf{P}_k)/K$ constrains the consistency of normals. 
The proposed robust SDF loss term $L_{{\rsdf}}$ regularizes the orientation of the deformed mesh, defined as 
 \begin{equation}
 \label{eq:robust_sdf}
 L_{\rsdf} = \frac{1}{K\cdot|\mathbf{Q}_s|}  \sum_{\mathbf{q}\in\mathbf{Q}_s} \sum_{k} \overline{\psi}_{\beta}^{k}(\mathbf{q}) \cdot |f_{\gamma}(s(\widehat{\mathbf{q}}^k)) - f_{\gamma} (\texttt{SDF}_{{\rm IMLS}}(\widehat{\mathbf{q}}^k, \mathbf{P}_k))|, 
\end{equation}  
where $\widehat{\mathbf{q}}^k$ is the deformed position of $\mathbf{q}$ according to Eq.~\eqref{eq:deformation}; $\overline{\psi}_{\beta}^{k}(\mathbf{q})=\frac{\psi_{\beta}^{k}(\mathbf{q})}{\sum_{k}\psi_{\beta}^{k}(\mathbf{q})}$ with ${\psi}_{\beta}^{k}(\mathbf{q})=\exp(-\beta\cdot|\texttt{SDF}_{{\rm IMLS}}(\widehat{\mathbf{q}}^k, \mathbf{P}_k)|^2)$ is to measure the confidence of each term. That is, when the deformed point $\widehat{\mathbf{q}}^k$ has a smaller value in the SDF approximated by the target point cloud, it is closer to the surface, and we assume that it is more accurate; $\mathbf{Q}_s$ is a subset randomly sampled from   
$\mathbf{Q}$. 
$f_{\gamma}(x)=\frac{1}{1+\exp({-\gamma\cdot x})}$ is used to scale the amplified SDF value to $(0, 1)$ because we pay more attention to whether the point $\widehat{\mathbf{q}}^k$ is inside or outside the surface. 

The loss term $L_{\smooth}$ ensures the smoothness of the deformation, defined as  
\begin{equation}
L_{\smooth} = \frac{1}{2K|\mathcal{E}|}\sum_{k}\sum_{l\in \mathcal{I}(k)} \sum_{(\mathbf{v}_i,\mathbf{v}_j)\in\mathcal{E}}\|({\mathbf{v}}_i^k-{\mathbf{v}}_i^l)-({\mathbf{v}}_j^k-{\mathbf{v}}_j^l)\|^2,
\end{equation}
where $\mathcal{E}$ is the edge set derived by face set $\mathbf{F}$. $\mathcal{I}(k) = \{k-1, k+1\}$. When $k=1$ or $k=K$, we only use one neighbor point cloud frame. Finally, it is crucial to ensure that the enhanced template surface maintains a reasonable shape throughout the joint optimization process. Thus, we constrain the deformation field at keyframe as close to the identity transformation as possible with a loss term $L_{\shape}$: 
\begin{equation}
L_{\shape} = \sum_{\mathbf{u}_r\in\mathbf{U}}\|[\xi_r^{k^*}, \mathbf{t}_r^{k^*}]\|^2, 
\end{equation}
where 
$[\xi_r^{k^*},\mathbf{t}_r^{k^*}]\in\mathbb{R}^6$ denotes the rotation matrix represented by Lie algebra format and translation vector of the keyframe.

\section{Experiments} 

\subsection{Experimental Settings} 
\mypara{Implementation details}
In our approach, we employed MLPs consisting of 5 linear layers with a feature dimension of 128 to implement $F_{\bf{\Phi}}(\cdot)$, $F_{\bf{\Xi}}(\cdot)$, and $F_{\bf{\Omega}}(\cdot)$. 
Training was conducted using the ADAM optimizer~\cite{kingma2014adam} with a learning rate of $1\times 10^{-4}$. 
The \textit{Supplementary Material} provides more details. 
We performed all experiments on a single NVIDIA A6000 GPU. 

\mypara{Datasets} 
We utilized three well-established benchmark datasets, namely DFA-UST~\cite{bogo2017dynamic}, DT4D~\cite{li20214dcomplete}, and AMA~\cite{vlasic2008articulated}, for our evaluation. From motion sequences, we selected 17 consecutive frames to form a set of input point clouds. Randomly sampling 5000 points from the ground truth mesh, we ensured diversity in our point cloud selection. Adopting the train and test data splits provided by~\cite{lei2022cadex}, we used 109 sets and 89 sets from the test data of the DFAUST dataset and DT4D-animal dataset, respectively. The AMA dataset comprises 10 motion sequences; for supervised learning comparison methods, we employed 1176 sets from 7 sequences as the train set and 32 sets of point clouds from the remaining 3 sequences ("crane", "march1" and "samba") as the test set.   To maintain consistency across sequences, we normalized the diagonal length of the bounding box in the first frame to 1 and centered it at the origin of the coordinates. Subsequent point clouds underwent transformations based on the scale and translation of the first frame. We utilized the algorithm in pymeshlab~\cite{pymeshlab} to estimate the normals of the point cloud and adjusted the normal directions to face outward.

\input{figs_scripts/compare_dfaust}

\begin{table*}[ht]
	\caption{
		Quantitative comparisons of different methods on three datasets. 
  } 
	\label{Tab:dfaust-dt4d-ama}
	\setlength{\tabcolsep}{5pt}
	\centering
  \resizebox{.8\textwidth}{!}{
		\begin{tabular}{ c | c| c  c  c  c  c }
			\toprule
            Dataset &{Method} &  CD ($\times 10^{-4}$) $\downarrow$ & NC $\uparrow$ & F-$0.5\%$ $\uparrow$ & F-$1\%$ $\uparrow$ & Corr. ($\times 10^{-2}$) $\downarrow$ \\
			\midrule
  \multirow{3}{*}{DFAUST~\cite{bogo2017dynamic}} & LPDC \cite{tang2021learning}& 2.430 & 0.929 & 0.299 & 0.633 & 1.46\\
   &Cadex \cite{lei2022cadex}&  1.062 & 0.941 & 0.519 & 0.823 & 1.28 \\
    &Ours & 0.688 & 0.953 & 0.894 & 0.985 & 1.02 \\\midrule 
   \multirow{2}{*}{DT4D~\cite{li20214dcomplete}} &Cadex \cite{lei2022cadex}& 22.377  & 0.868 & 0.269 & 0.532 & 4.99 \\
    &Ours & 1.036 & 0.933 & 0.582 & 0.870 & 3.23\\\midrule
    \multirow{3}{*}{AMA~\cite{vlasic2008articulated}} &  LPDC \cite{tang2021learning}& 108.822  & 0.666 & 0.045 & 0.101 & 14.1\\
   & Cadex \cite{lei2022cadex}& 71.041 &0.663 & 0.052& 0.119 & 13.6 \\
   & Ours & 0.320 & 0.918 &0.683 & 0.943 & 4.44 \\
			\bottomrule
		\end{tabular}
  }
\end{table*}

\input{figs_scripts/compare_ama}

\mypara{Methods under comparison} We conducted a comparative analysis involving our proposed method and two cutting-edge techniques specialized in dynamic surface reconstruction from sequences of point clouds: LPDC~\cite{tang2021learning} and Cadex~\cite{lei2022cadex}. \textit{Note} that both Cadex and LPDC necessitate \textit{ground-truth occupancy values} and \textit{ground-truth correspondences} for supervision during training. For Cadex, we employed the officially released pre-trained model provided by the authors for testing on DFAUST and DT4D datasets. As for LPDC, the authors exclusively performed experiments on the DFAUST dataset and released the pre-trained model. To ensure a comprehensive comparison, we added the comparison on the AMA dataset, a human body dataset characterized by \textit{significantly larger} deformation and movement ranges compared to DFAUST. Furthermore, we retrained Cadex and LPDC on the AMA dataset.

\mypara{Evaluation metrics} 
Following previous works~\cite{mescheder2019occupancy, lei2022cadex, peng2021shape}, we scaled the test data to its original size and quantitatively assessed various methods using $\ell_2$-norm-based Chamfer Distance (CD), Normal Consistency (NC), F-score with thresholds of $0.5\%$ (F-$0.5\%$) and $1\%$ (F-$1\%$), and Correspondences Error (Corr.) to evaluate different methods quantitatively. 
For the visual results, we assigned the same colors to corresponding vertices to indicate the accuracy of learned correspondences by different methods. 
In addition, we also show the error map to visualize the pointwise distance error from the ground-truth surface. 

\subsection{Comparisons with State-of-the-Art Methods}

\mypara{Results of clean and complete data}
Table~\ref{Tab:dfaust-dt4d-ama} shows the comparison of our method against state-of-the-art methods on the three datasets. The results demonstrate that our method attains the highest accuracy. The visual comparisons in Figs.~\ref{fig:dfaust} and \ref{fig:ama} further validate the significant superiority of our method. Notably, Cadex and LPDC exhibit notably poor performance on the AMA dataset. This discrepancy could be attributed to the smaller data volume of the AMA dataset compared to DFAUST and DT4D, shedding light on the limitations of these supervised learning methods. We also refer readers to the \textit{Supplementary Material} and \textit{Video Demo} for more visual results.

\begin{figure}[htb]
    \centering
    \includegraphics[width=0.9\columnwidth]{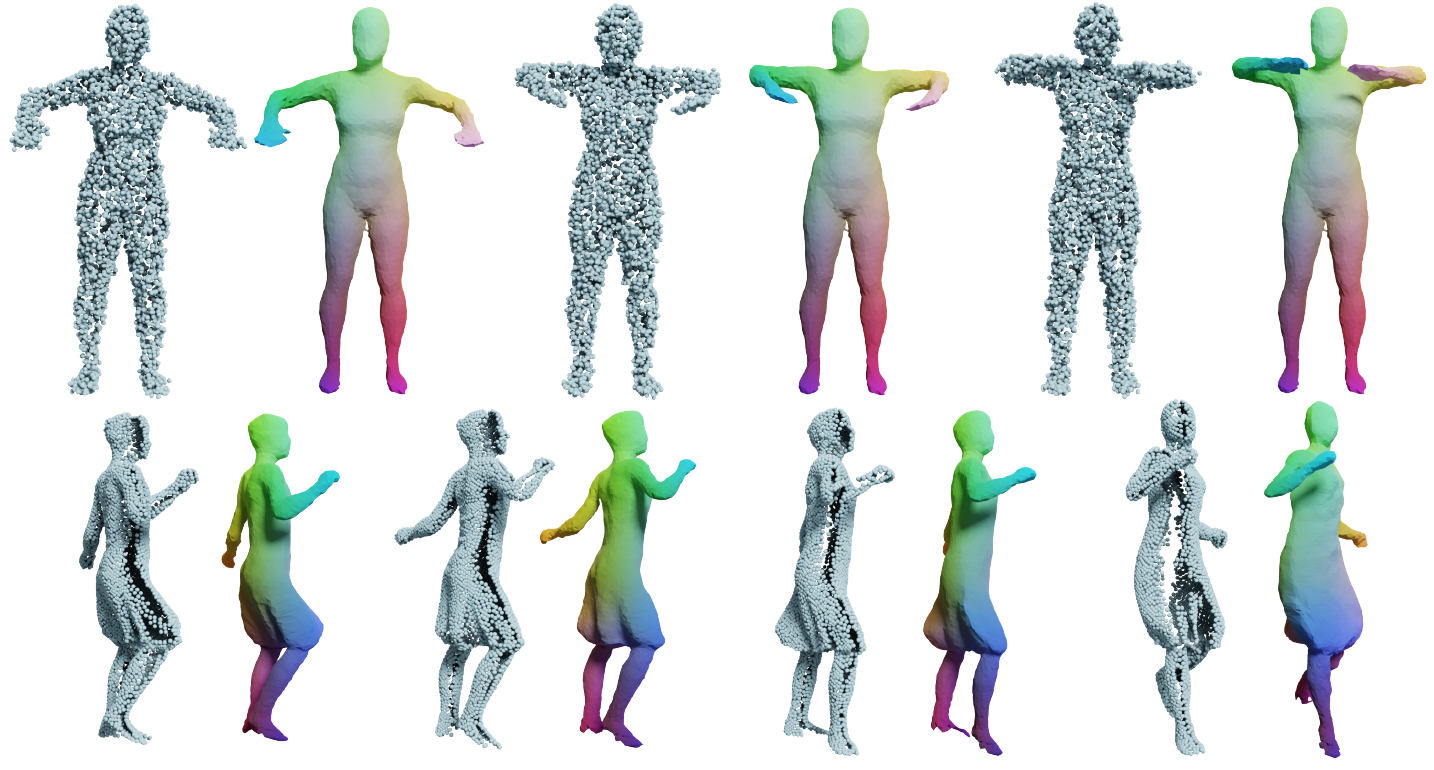}  
    \caption{Results by our method on noisy data (top) and partially missing data (bottom). }
    \label{fig:noise-partial} 
\end{figure}

\mypara{Results of noisy and partially missing data} We also evaluate the performance of our method on noisy data. We add $0.5\%$ Gaussian noise for each point in a point cloud sequence in DFAUST dataset~\cite{bogo2017dynamic} and show the results in Fig.~\ref{fig:noise-partial}. Since point cloud data collected by real-depth cameras often contain missing parts, we tested the situation when there are some holes in the surface. For a sequence in "swing" motion sequences of AMA dataset~\cite{vlasic2008articulated}, we placed two virtual cameras in front of and behind the mesh model for simulated acquisition and merged the point clouds scanned by the two cameras. Fig.~\ref{fig:noise-partial} shows the performance of our method and the variants. We can see that our method can reconstruct partially missing meshes. More results are presented in the \textit{Supplementary Material}.

\subsection{Ablation Study}

\begin{figure}[t]  
\centering 
{
\begin{tikzpicture}[]

\node[] (a) at (5,3.) {\includegraphics[width=0.95\textwidth]{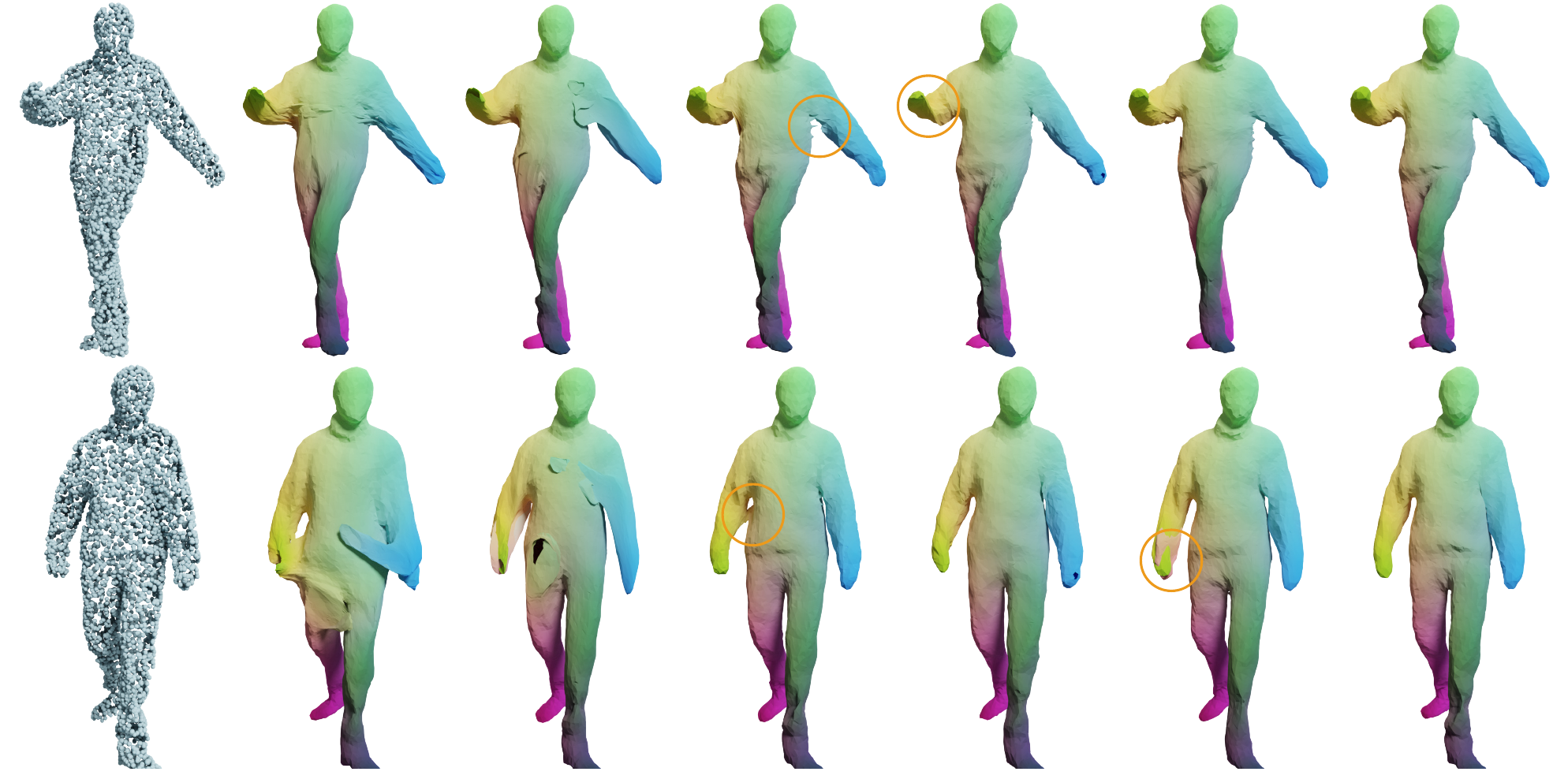}};

\node[] (a) at (0,0) { (a)};

\node[] (b) at (1.7,0) { (b)};

\node[] (c) at (3.4,0) { (c)};

\node[] (d) at (5.03,0) { (d)};

\node[] (e) at (6.7,0) { (e)};

\node[] (f) at (8.35,0) { (f)};

\node[] (g) at (10,0) { (g)};
\end{tikzpicture}
}
\caption{Comparison of visual results by different variants of the deformation field in our method. (a) Input; (b) Point-wise deformation based on an MLP; (c) Fixed blending weight formula; (d) Fixed control point positions; (e) $|\mathbf{U}|=10$; (f) $|\mathbf{U}|=100$; (g) Ours.}
\label{fig:ablas-deformation}
\end{figure}

\begin{figure}[bt] 
\centering 
{
\begin{tikzpicture}[]

\node[] (a) at (5,2) {\includegraphics[width=0.95\textwidth]{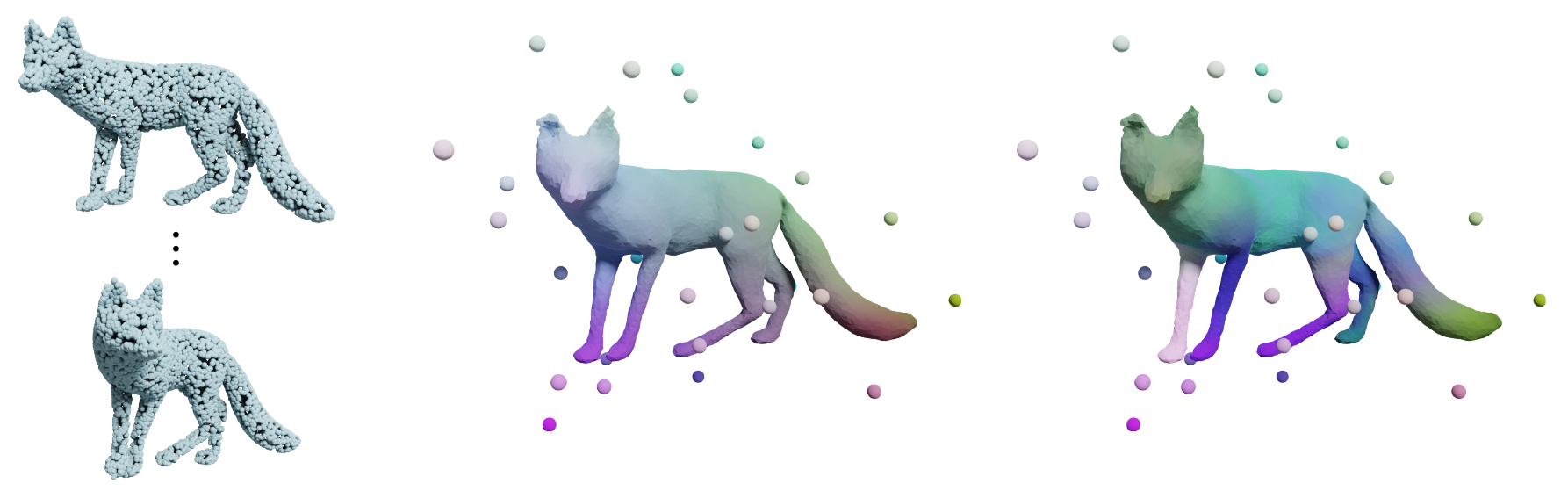}};

\node[] (a) at (0.4,0) { (a) Input};

\node[] (b) at (4.5,0) { (b) Computed by Eq.~\eqref{eq:init_blending_weight}}; 

\node[] (c) at (8.6,0) { (c) Ours};

\end{tikzpicture}
} 
\caption{Visual learnable control points and blending weights. We assigned a color $\mathbf{c}_r$ to each control point $\mathbf{u}_r$ based on its spatial position, and obtained the color for each $\mathbf{\overline{v}}$ by $\sum_{r=1}^{|\mathbf{U}|}\omega(\mathbf{\overline{v}},\mathbf{u}_r)\cdot \mathbf{c}_r$. Closer colors indicate more similar deformation fields. 
}
\label{fig:vis_node_weight}
\end{figure}

In this section, we conducted thorough ablation studies targeting key components of our framework on the AMA dataset, where we took 1 set from each sequence and 10 sets in total for data diversity. We focused on the following aspects: 

\begin{table*}[t]
	\caption{
		 Ablation studies of our method on the AMA dataset~\cite{vlasic2008articulated}. The best and second best results are highlighted in \textbf{bold} and \underline{underlined}.} 
	\label{Tab:ablations}
	\setlength{\tabcolsep}{3pt}
	\centering
 \resizebox{.8\textwidth}{!}{
		\begin{tabular}{ c | c  c  c  c  c }
			\toprule
             {Variants} &  CD($\times 10^{-4}$) $\downarrow$ & NC $\uparrow$ & F-$0.5\%$ $\uparrow$ & F-$1\%$ $\uparrow$ & Corr. ($\times 10^{-2}$) $\downarrow$ \\
			\midrule
    MLP-Deform & 0.412 &  0.909 & 0.618 & 0.917  & 5.91 \\
    w. $\tilde{\omega}(\mathbf{\overline{v}},\mathbf{u}_r)$  & 0.475 & 0.911 & 0.564 & 0.890  & 3.26\\
    Fixed $\{\mathbf{u}_r\}$  & 0.372& 0.911 & 0.637 & 0.927  & 3.23 \\
   \midrule 
   w.o. $L_{\rsdf}$  & 0.330 & 0.918 & 0.649 & 0.937  & 3.14\\
   w.o. $L_{\normal}$  & 0.656 & 0.900 & 0.582 & 0.884  & 3.30 \\
   w.o. $L_{\smooth}$  & 0.378 & 0.914 & 0.616 & 0.923  & 3.30 \\
   w.o. $L_{\shape}$  & 0.335 & 0.919 & 0.636 & 0.936  & \underline{2.91} \\
   \midrule 
   $|\mathbf{U}|=10$ & 0.437 & 0.910 & 0.594 & 0.910 &  3.30 \\
   $|\mathbf{U}|=30$ (\textbf{Ours}) & \underline{0.324} & \underline{0.919} & \underline{0.649} & \underline{0.939} &  \textbf{2.89} \\
   $|\mathbf{U}|=100$ & \textbf{0.315} & \textbf{0.919} & \textbf{0.651} & \textbf{0.943} &  3.16 \\
			\bottomrule
		\end{tabular}
  }
\end{table*}
\mypara{1) Learnable deformation field}  
To evaluate the effectiveness of the proposed deformation field, we conducted experiments by maintaining the other components and substituting the deformation field: 
\begin{itemize}
\item \textit{Pointwise deformation based on an MLP}: For any point $\overline{\mathbf{v}}\in\mathbb{R}^3$ in the template surface and frame index $k$, we established a deformation function $F_{\bf{\Psi}}(\Gamma([\overline{\mathbf{v}}||k]))$ and obtained the deformed position $\mathbf{v}^k=\overline{\mathbf{v}}+F_{\bf{\Psi}}(\Gamma([\overline{\mathbf{v}}||k]))$. Here, $F_{\bf{\Psi}}(\cdot)$ is an MLP with 5 linear layers and a feature dimension of 128.
\item \textit{Fixed blending weight formula}: 
We replaced the learnable blending weights in Eq.~\eqref{eq:deformation} by setting it to $\tilde{\omega}(\mathbf{\overline{v}},\mathbf{u}_r)$ in Eq.~\eqref{eq:init_blending_weight} throughout the learning process.
\item \textit{Fixed control point positions}: We maintained the positions ${\mathbf{u}_r}$ of the control points at their initial values. 
\end{itemize}
From Tab.~\ref{Tab:ablations} and Fig.~\ref{fig:ablas-deformation}, 
the three variants produce worse quantitative and visual results than Ours, verifying the advantage of our proposed deformation field.  Fig.~\ref{fig:vis_node_weight} also visualizes the learned weights via Eq. \eqref{eq:learnable weights} and those computed via Eq.~\eqref{eq:init_blending_weight} for blending, 
where we can see that the learned weights can capture the motion information and well handle the deformations for the parts with close spatial locations but inconsistent motions.

\mypara{2) Size of $|\mathbf{U}|$} From Tab.~\ref{Tab:ablations} and Fig.~\ref{fig:ablas-deformation}, where we set different numbers of control points i.e., 10, 30, and 100, we can observe that too few control points result in insufficient degrees of freedom and suboptimal alignment, while too many control points can weaken local consistency. We set $|\mathbf{U}|=30$ as a balanced trade-off.

\mypara{3) Loss functions during temporal reconstruction}
To assess the effectiveness of various loss terms used in the temporal reconstruction stage, we individually excluded the robust SDF loss $L_{\rsdf}$, deformation smooth loss $L_{\smooth}$, and shape preservation loss $L_{\shape}$, while keeping other aspects unchanged. 
The results presented in 
Tab.~\ref{Tab:ablations} demonstrate their necessity and contributions. 

\mypara{4) Keyframe selection, template surface learning and arbitrary length of input sequences} The performances and discussions can be found in the \textit{Supplementary Material}.
\section{Conclusion and Discussion}
We introduced a new unsupervised temporally-consistent dynamic surface reconstruction framework for time-varying point cloud sequences. Technically, we integrated template surface representation based on the deformable tetrahedron and a learnable deformation field. We first proposed a coarse-to-fine learning strategy for constructing the template surface. Then we designed a learnable deformation field using the learnable control points and blending weights for temporally reconstruction.  
We jointly optimized the enhanced template surface and the learnable deformation field. Experiments demonstrate that our proposed method performs even much better than SOTA supervised learning methods. 

\begin{wrapfigure}[10]{r}{0.4\textwidth}
\includegraphics[width=\linewidth]{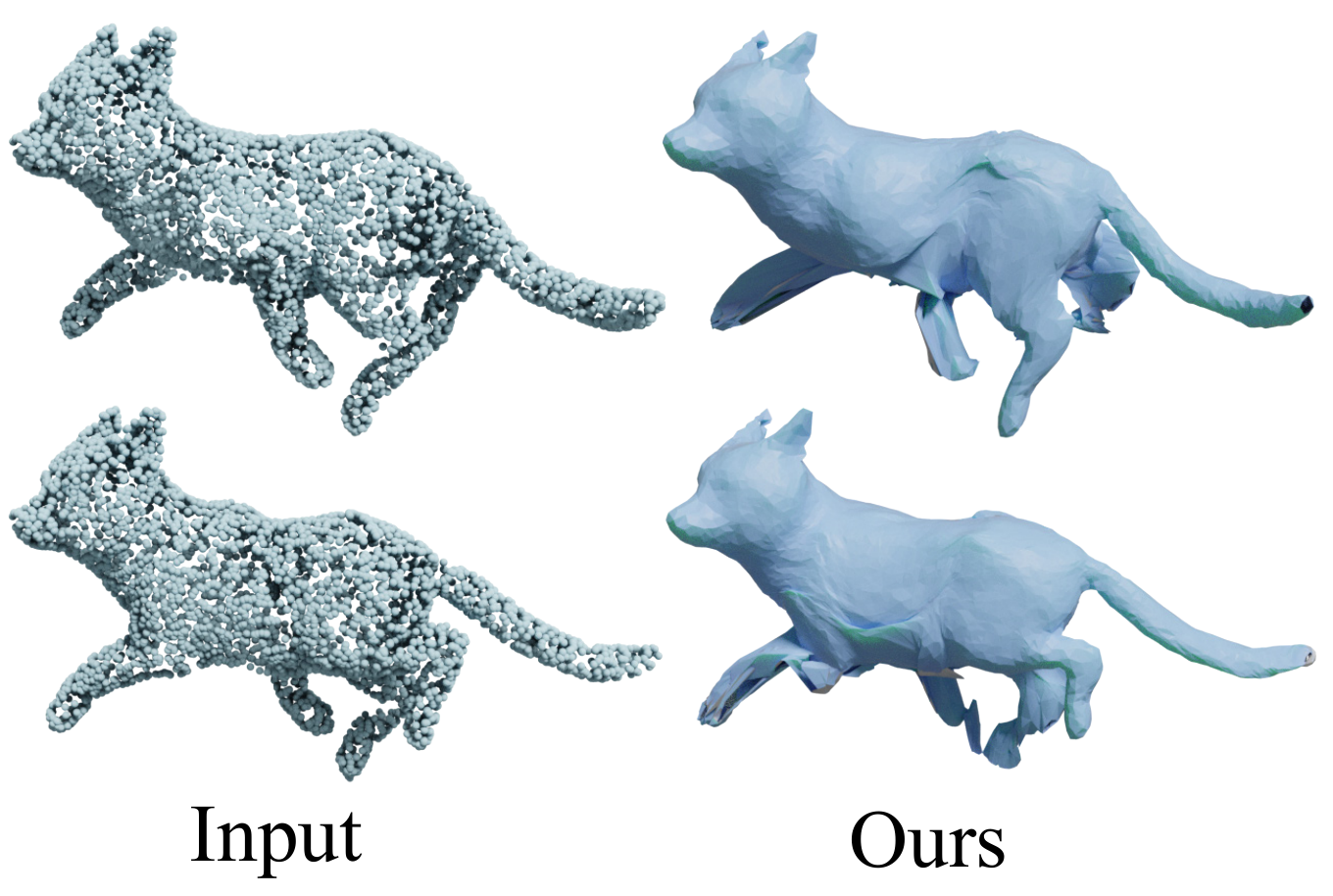}
\end{wrapfigure}
Although our method has achieved good performance on many examples, it may encounter limitations in cases where the point cloud in certain areas is sparse or the local structure is elongated but exhibits significant non-rigid deformations (as depicted on the right). This is due to our utilization of Chamfer distance as the alignment supervision, which restricts the establishment of correspondences between adjacent frames solely based on spatial proximity. Incorporating more robust metrics for shape alignment would enhance the effectiveness of our approach.

\section*{Acknowledgement} This project was supported by the Hong Kong Research Grants Council under Grants 11219324, 11219422, and 11202320.

% ---- Bibliography ----
\bibliographystyle{splncs04}
\bibliography{main}

\clearpage
\appendix 
\vspace{1cm}
\section*{\centering \large Supplementary Material}
\vspace{0.3cm}

\section{Supplement of ablation study}  

In this section, we further discuss the effectiveness of some components of our method, as well as the robustness of the input data.

\subsubsection{Keyframe selection and template surface learning.}
Compared to directly setting the middle frame as a keyframe, Fig.~\ref{fig:ablas-template} shows that the selected keyframe by our strategy is closer to the average shape. After learning the template surface, we also show the results of only using the fine stage (Fig.~\ref{fig:ablas-template} (b)) and using the coarse-to-fine stage (Fig.~\ref{fig:ablas-template} (c)). Based on Fig.~\ref{fig:ablas-template} (c), we further show the enhanced template surface after the temporal reconstruction process. 
\input{figs_scripts/ablas_template}

\subsubsection{Arbitrarily long point cloud sequence.}
Our method is theoretically {independent} of the sequence length. We experimented with sequences of various lengths on "50007$\_$shake$\_$shoulders" sequence from DFAUST. As shown in Tab.~\ref{Tab:long_seq} below, our method can effectively handle sequences of varying lengths. When the sequence is relatively long, the reconstruction quality decreases slightly due to the diversity of deformations. 
\begin{table}[ht]
	\caption{
  Results of our method on sequences with various lengths.} 
	\label{Tab:long_seq}
	\setlength{\tabcolsep}{1 pt}
	\centering
 \resizebox{0.85\columnwidth}{!}{
		\begin{tabular}{ c | c  c  c  c  c }
			\toprule
             {Sequence length} &  CD($\times 10^{-4}$) $\downarrow$ & NC $\uparrow$ & F-$0.5\%$ $\uparrow$ & F-$1\%$ $\uparrow$  & Corr. ($\times 10^{-2}$) $\downarrow$ \\
			\midrule
   10  &  0.154  &  0.957  & 0.866 &  0.992  & 0.53 \\
   30  &  0.138 &  0.954 & 0.876 &  0.994  & 0.97  \\
   50  &  0.167 &  0.955 & 0.831 &  0.989  & 1.46 \\
   80  &  0.265 &  0.946 & 0.704 &  0.960  & 1.72 \\
			\bottomrule
		\end{tabular}
  }
\end{table}

\subsubsection{Various data qualities.} 
We experimented with point clouds of different qualities on the AMA dataset:  (\textbf{1}) we randomly sampled $N_k$
points from the ground truth mesh respectively as input of our method; and 
(\textbf{2}) we randomly sampled $5000$ points from ground truth mesh, randomly selected $10$ points from them, and deleted the $h_k$ nearest neighbor points of these points to construct point clouds with holes. As listed in Tab.~\ref{Tab:data_quality}, our method shows {robustness} to variations in {point cloud density and holes}. We also showcase the visualization of constructed point clouds with various qualities in Fig.~\ref{fig:supp-data-qualities}.

\begin{table}[hbt]
\footnotesize
	\caption{
  {Results of our method on point clouds with various qualities. }} 
	\label{Tab:data_quality}
	\setlength{\tabcolsep}{1pt}
	\centering
 \resizebox{0.85\columnwidth}{!}{
		\begin{tabular}{ c  c | c  c  c  c  c }
			%\Xhline{1pt}
			\toprule
              \multicolumn{2}{c|}{Input Data} &  CD($\times 10^{-4}$) $\downarrow$ & NC $\uparrow$ & F-$0.5\%$ $\uparrow$ & F-$1\%$ $\uparrow$  & Corr. ($\times 10^{-2}$) $\downarrow$ \\
			\midrule
   \multirow{3}{*}{(\textbf{1})} & $N_k=2000$  & 0.451  & 0.906  & 0.563 &  0.899  & 3.54 \\
   &$N_k=3000$ & 0.402  & 0.911  & 0.610 & 0.916   & 3.46  \\
   & $N_k=4000$  & 0.381  & 0.913  & 0.597 &  0.921  & 3.41 \\
   \midrule
   \multirow{3}{*}{(\textbf{2})} & $h_k=10$  & 0.336  & 0.918  & 0.634 & 0.935   & 2.88  \\
   &$h_k=20$  & 0.366 & 0.914  & 0.617 & 0.926   & 3.57 \\
    &$h_k=30$  & 0.405  & 0.911  & 0.587 & 0.912   & 3.50 \\
   \midrule
   \multicolumn{2}{c|}{GT-normal} &  0.318 &  0.918 & 0.656 &  0.940  & 3.10 \\
   \midrule
   \multicolumn{2}{c|}{\textbf{Ours}} &  0.324 &  0.919 & 0.649 &  0.939  & 2.89 \\
			\bottomrule
		\end{tabular}
  }
\end{table}
\input{figs_scripts/supp_data_quality}

\subsubsection{Dependence on the normal direction.}  In all experiments, we {did not use any ground-truth} normals. Instead, we first use the algorithm in PyMeshLab~\cite{pymeshlab} to {estimate} the normals of input point clouds. We then adjust the normal directions to ensure that the majority of normals point outward. Besides, we also replaced estimated normals with ground-truth normals in our method. As listed in Tab.~\ref{Tab:data_quality}, the performance is comparable to {Ours},
demonstrating the {robustness} of our method to normals of input point clouds.

\subsubsection{Visualization of omitting loss terms.}
\begin{figure}[hbt] \small 
\centering 
{
\begin{tikzpicture}[]

\node[] (a) at (5,3.6) {\includegraphics[width=0.95\textwidth]{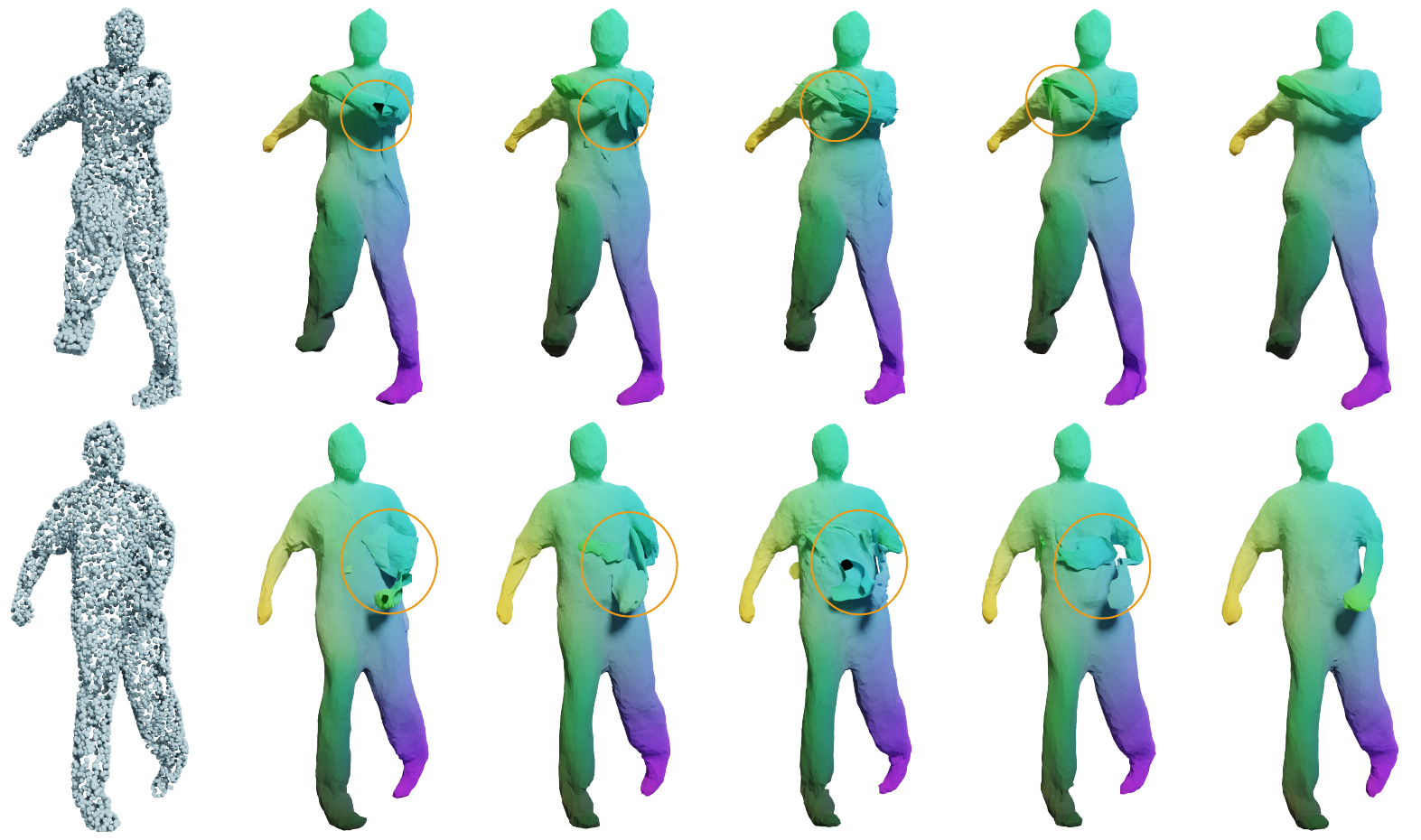}};

\node[] (a) at (0.1,0) {\small (a)};

\node[] (b) at (2.1,0) {\small (b)};

\node[] (c) at (4.1,0) {\small (c)};

\node[] (d) at (6.1,0) {\small (d)};

\node[] (e) at (8.1,0) {\small (e)};

\node[] (f) at (10.1,0) {\small (f)};

\end{tikzpicture}
}
\caption{Comparison of visual results by different variants of excluding a certain loss in our method. (a) Input; (b) w.o. $L_{\rsdf}$; (c) w.o. $L_{\normal}$; (d) w.o. $L_{\smooth}$; (e) w.o. $L_{\shape}$; f) Ours.
}
\label{fig:ablas-loss}
\end{figure}

Additionally, Fig.~\ref{fig:ablas-loss} illustrates the visual results of our method when different loss terms are omitted. We can see that our method, incorporating all loss terms, achieves the best results.

\input{figs_scripts/supp_ama}

\section{More visual results}
We present more visual results in this section and showcase the complete motion sequences in the \textit{Video Demo}.
\subsubsection{Comparisons with state-of-the-art methods.}
We present additional visual results for comparisons with LPDC~\cite{tang2021learning} and Cadex~\cite{lei2022cadex} on the AMA dataset~\cite{vlasic2008articulated} (Fig.~\ref{fig:supp-ama}), DT4D dataset~\cite{li20214dcomplete} (Fig.~\ref{fig:supp-dt4d}) and DFAUST dataset~\cite{bogo2017dynamic} (Fig.~\ref{fig:supp-dfaust}). It is evident that our method outperforms other approaches.

\input{figs_scripts/supp_dt4d}

\input{figs_scripts/supp_dfaust}

\subsubsection{Performance on noisy data and partially missing data.} 
\begin{figure}[h]
    \centering
    \includegraphics[width=0.95\columnwidth]{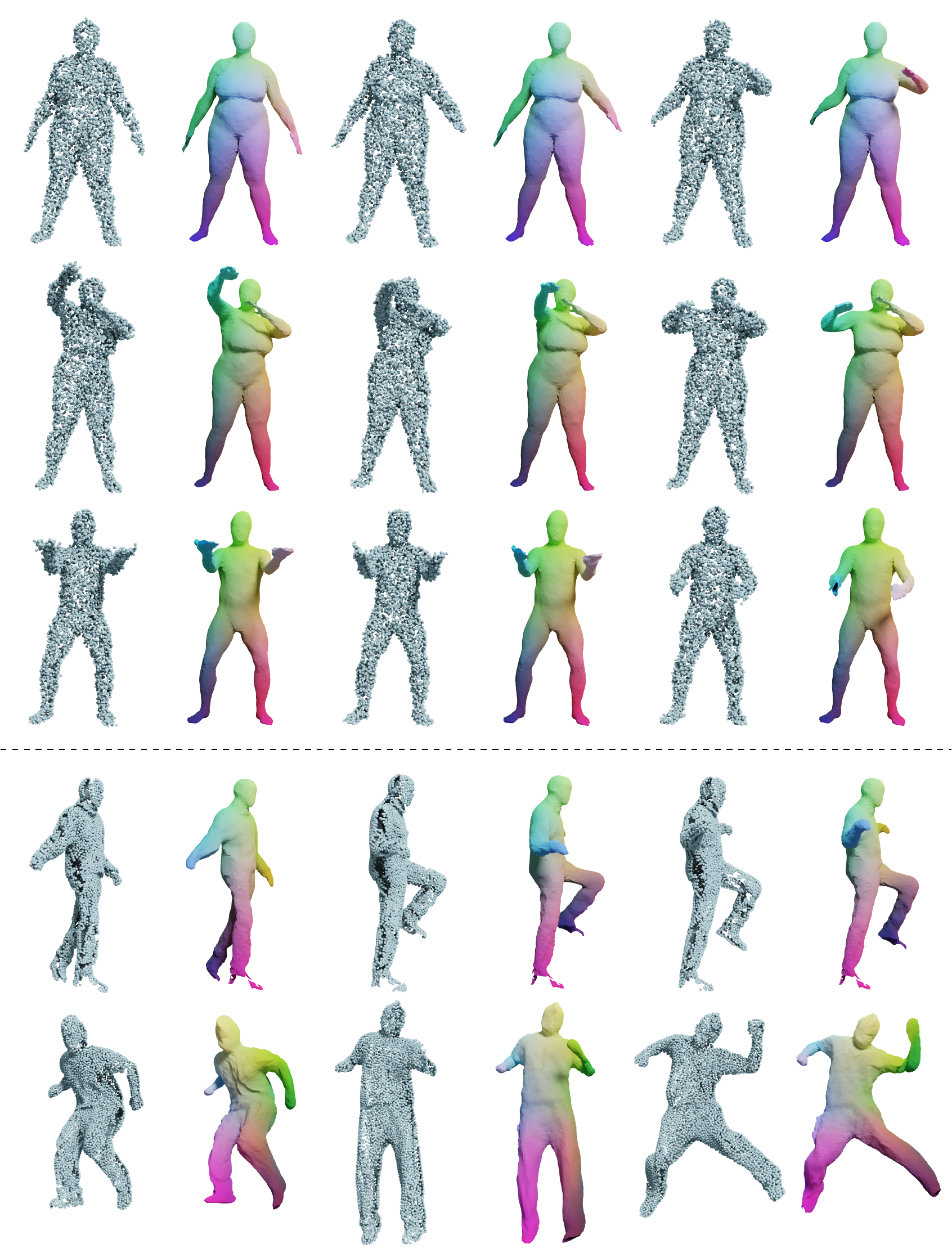} 
    \caption{Visual results of our method on constructed noisy data from the DFAUST dataset~\cite{bogo2017dynamic} (Top) and partially missing data from AMA dataset~\cite{vlasic2008articulated} (Bottom).}
    \label{fig:supp-noise} 
\end{figure}

Furthermore, in Fig.~\ref{fig:supp-noise}, we showcase more visual results of our method applied to noisy data constructed from the DFAUST dataset~\cite{bogo2017dynamic} and partially missing data constructed from AMA dataset~\cite{vlasic2008articulated}, respectively.

\section{Technical Details}
In this section, we introduce the parameter settings and explain some definitions in the loss functions and evaluation metrics mentioned in the paper.

\subsubsection{Parameter settings.}
Reconstructing temporally consistent dynamic surfaces from point cloud sequences is challenging, especially for our unsupervised learning scenario. Therefore, we need multiple hyperparameters associated with the regularization terms for shrinking the large solution space of our unsupervised pipeline, we have uniformly set them to default values in all experiments and found that they generally work well for most examples, minimizing the need for additional parameter adjustments for new datasets. By default, in the calculation of the loss function (8), we set the number of the points for both $\widetilde{\mathbf{V}}_k$ in Eq. (9) and $\mathbf{Q}_s$ in Eq. (10) to be $10^4$. We set $\eta=0.1$ in Eq.~(7), $\alpha=5.56$ in Eq.~(9) and $\beta=50, \gamma=10^2$ in Eq.~(10).  
The default weights setting in the loss function are listed in Tab.~\ref{Tab:parameter}. In addition, we also showcase the number of iterations and training time in this table.

\begin{table*}[ht]
	\caption{ Default parameter settings and runtime. For the learning template surface, we listed iterations and running time for the coarse/fine stage.} 
	\label{Tab:parameter}
	\setlength{\tabcolsep}{3pt}
	\centering
  \resizebox{.95\textwidth}{!}{
		\begin{tabular}{ c  c  c c c  | c c c  c  c c c }
  \toprule
    \multicolumn{5}{c|}{Learning template surface} & \multicolumn{7}{c}{Temporal reconstruction}\\
    \cmidrule{1-5} \cmidrule{6-12} 
    $\tilde{w}_1$ & $\tilde{w}_2$ & $\tilde{w}_3$ & $\#$Iters & Time & $w_1$ & $w_2$ & $w_3$ & $w_4$ & $w_5$ & $\#$Iters & Time \\	
    \midrule
   $5\times 10^2$ & $10^{-3}$ & $50$ & $10^{3}$ / $5\times 10^3$ & $10$ s/ $2$ mins & $5\times 10^2$ & $10^{-3}$ & $10^2$ & $10^3$ & $1$ & $10^{4}$ & 30 mins\\ 
			\bottomrule
		\end{tabular}
  }
\end{table*}

\subsubsection{The details of loss functions.}
The Chamfer Distance between $\mathbf{X}$ and $\mathbf{Y}$ is defined as 
\begin{equation*}
\label{eq:chamfer-distance}
\texttt{CD}_{\ell_p}(\mathbf{X}, \mathbf{Y}) = \frac{1}{|\widetilde{\mathbf{X}}|}\sum_{\mathbf{x}\in\widetilde{\mathbf{X}}} \left\|\mathbf{x}-\hat{\mathbf{y}}\right\|_{\ell_p} + \frac{1}{|\widetilde{\mathbf{Y}}|}\sum_{\mathbf{y}\in\widetilde{\mathbf{Y}}} \left\|\mathbf{y}-\hat{\mathbf{x}}\right\|_{\ell_p},
\end{equation*}
where $p=1,2$. $\hat{\mathbf{y}}\in\widetilde{\mathbf{Y}}$ and $\hat{\mathbf{x}}\in\widetilde{\mathbf{X}}$ are the closest points for the $\mathbf{x}$ and $\mathbf{y}$ respectively. When $\mathbf{X} (\text{or } \mathbf{Y})$ is a point set, $\widetilde{\mathbf{X}}=\mathbf{X} (\text{or }  \widetilde{\mathbf{Y}}=\mathbf{Y})$. When $\mathbf{X} (\text{or } \mathbf{Y})$ represents a mesh, $\widetilde{\mathbf{X}} (\text{or }  \widetilde{\mathbf{Y}})$ denotes the sampling point on the mesh. In the learning phase, we set the number of sampling points to $10^{4}$, and during the evaluation, we set the number of sampling points to $10^{5}$. 
 
In Eqs. (3) and (8) of the paper, the normal consistency of $\mathbf{X}$ and ${\mathbf{Y}}$ is defined as 
\begin{equation*}
\label{eq:normal-consistent}
\texttt{NC}_{\ell_1}(\mathbf{X}, \mathbf{Y}) = \frac{1}{|\mathbf{N}_x|}\sum_{\mathbf{n}_x\in\mathbf{N}_x} \left|1-|\langle\mathbf{n}_x, \mathbf{n}_{\hat{y}}\rangle|\right| + \frac{1}{|\mathbf{N}_y|}\sum_{\mathbf{n}_y\in\mathbf{N}_y} \left|1-|\langle \mathbf{n}_y,{\mathbf{n}}_{\hat{x}}\rangle|\right|,
\end{equation*}
where $\mathbf{N}_x$ and $\mathbf{N}_y$ are the normal sets of $\widetilde{\mathbf{X}}$ and $\widetilde{\mathbf{Y}}$, respectively, $\mathbf{n}_{\hat{y}}\in\mathbf{N}_y$ and $\mathbf{n}_{\hat{x}}\in\mathbf{N}_x$ are the corresponding normals of $\hat{\mathbf{y}}$ and $\hat{\mathbf{x}}$ respectively. $\langle \cdot , \cdot \rangle$ is the inner product of two vectors. 

In Eq.~(4), the SDF value $\texttt{SDF}_{{\rm IMLS}}(\mathbf{q}, \mathbf{P}_{k^*})$ at $\mathbf{q}$ approximated through implicit moving least-squares is defined as 
\begin{equation*}
\texttt{SDF}_{{\rm IMLS}}(\mathbf{q}, \mathbf{P}_{k^*}) = \frac{\sum_{\mathbf{p}_j\in\mathcal{N}(\mathbf{q})}\theta(\|\mathbf{q}-\mathbf{p}_j\|,{\zeta})\cdot \langle\mathbf{q}-\mathbf{p}_j, \mathbf{n}_j\rangle}{\sum_{\mathbf{p}_j\in\mathcal{N}(\mathbf{q})}\theta(\|\mathbf{q}-\mathbf{p}_j\|, \zeta)}, 
\end{equation*}
where $\mathcal{N}(\mathbf{q})$ denotes the set of nearest points in $\mathbf{P}_{k^*}$. $\theta(d,\zeta) = \exp(-d^2/\zeta^2)$ is the weight of each point in $\mathcal{N}(\mathbf{q})$. Here we set $|\mathcal{N}(\mathbf{q})|=10$ and $\zeta=0.1$ by default.

\subsubsection{Evaluation metrics.}
Following~\cite{mescheder2019occupancy,peng2021shape,ren2023geoudf}, the F-score to evaluate the reconstruction accuracy is defined as:
\begin{equation*}
\textrm{F-score}(\mathcal{M}_k,\mathcal{M}_k^{\textrm{GT}},\epsilon)=\frac{2\cdot \textrm{Recall}\cdot\textrm{Precision}}{\textrm{Recall}+\textrm{Precision}},
\end{equation*}
where 
\[
\textrm{Recall}(\mathcal{M}_k,\mathcal{M}_k^{\textrm{GT}},\epsilon) = \left|\left\{\mathbf{p}_1\in\widetilde{\mathbf{P}}_k, \textrm{s.t.} \min_{\mathbf{p}_2\in\widetilde{\mathbf{P}}_k^{\textrm{GT}}}\|\mathbf{p}_1-\mathbf{p}_2\|<\epsilon\right\}\right|,
\]
\[
\textrm{Precision}(\mathcal{M}_k,\mathcal{M}_k^{\textrm{GT}},\epsilon) = \left|\left\{\mathbf{p}_2\in\widetilde{\mathbf{P}}_k^{\textrm{GT}}, \textrm{s.t.} \min_{\mathbf{p}_1\in\widetilde{\mathbf{P}}_k}\|\mathbf{p}_1-\mathbf{p}_2\|<\epsilon\right\}\right|.
\]
Here $\mathcal{M}_{k}^{\textrm{GT}}$ is the $k$-th ground-truth mesh,
$\widetilde{\mathbf{P}}_k (\widetilde{\mathbf{P}}_k^{\textrm{GT}})$  
is the point set of the randomly sampling $10^5$ points from $\mathcal{M}_k (\mathcal{M}_{k}^{\textrm{GT}})$. We set $\epsilon$ to $0.5\%$ or $1\%$, and compute mean value of $\textrm{F-score}(\mathcal{M}_k,\mathcal{M}_k^{\textrm{GT}},\epsilon)$ ($k=1,2,...,K$) to get F-$0.5\%$ or F-$1\%$.

We denote $\{\mathbf{P}^{\textrm{GT}}(k)\}_{k}$ as the temporally consistent corresponding points on the ground-truth surfaces. We set $|\mathbf{P}^{\textrm{GT}}(k)|=10^{5}$. In the first frame, we construct the index set of nearest points:
\[
\texttt{Index} = \{\rho_i | \mathbf{v}_{\rho_i}=\mathop{\arg\min}_{\mathbf{v}\in\mathbf{V}_1}\|\mathbf{v}-\mathbf{p}_i^{\textrm{GT}}(1)\|\},
\]
where $\mathbf{V}_1$ is the vertex set of $1$-st reconstructed mesh $\mathcal{M}_1$, $\mathbf{p}_i^{\textrm{GT}}(1)\in\mathbf{P}^{\textrm{GT}}(1)$. Following~\cite{niemeyer2019occupancy,lei2022cadex}, the Correspondences Error is defined as 
\[
\textrm{Corr.} = \frac{1}{|\texttt{Index}|\cdot K}\sum_{k=1}^{K}\sum_{\rho_i\in\texttt{Index}}\|\mathbf{v}_{\rho_i}^{k}-\mathbf{p}_i^{\textrm{GT}}(k)\|.
\]

\end{document}